\makeatletter\def\graphicscache@inhibit{true}\makeatother

\documentclass[letterpaper, 10 pt, conference]{ieeeconf}  %

\IEEEoverridecommandlockouts                              %

\usepackage[utf8]{inputenc}
\usepackage[T1]{fontenc}
\usepackage{textcomp}
\usepackage{graphicx} %
\usepackage{todonotes}
\usepackage{pgfplots}
\pgfplotsset{compat=newest}
\usepackage{subfig}
\usepackage{multirow}
\usepackage{tabularx}
\usepackage[hidelinks]{hyperref}
\usepackage{array}
\usepackage{float}
\usepackage{graphbox}
\usepackage{tikz,tikz-3dplot}
\usetikzlibrary{arrows,arrows.meta,automata,backgrounds,calc,chains,%
decorations.markings,decorations.pathreplacing,decorations.pathmorphing,%
matrix,positioning,shapes,shapes.geometric,shapes.symbols,spy,trees,tikzmark}
\usepackage{balance}
\usepackage[binary-units=true,product-units=single,per-mode=symbol,range-units=single,detect-all]{siunitx}
\DeclareSIUnit\pixel{px}
\usepackage{amsmath} %
\usepackage{amssymb}  %

\usepackage{graphicscache}

\DeclareMathOperator{\atantwo}{atan2}

\newcommand{\reffig}[1]{Fig.~\ref{#1}}

\newcommand{\refsec}[1]{Sec.~\ref{#1}}

\newcommand{\etal}{et al.~}
\newcommand{\citep}[1]{(\cite{#1})}

\newcommand{\ie}{i.e.,\ }
\newcommand{\eg}{e.g.,\ }
\newcommand{\cf}{cf.\ }

\usepackage[font=small]{caption}

\makeatletter
\tikzset{
from/.style args={#1 to #2}{%
        above right={0cm of #1},%
        /utils/exec=\pgfpointdiff
            {\tikz@scan@one@point\pgfutil@firstofone(#1)\relax}
            {\tikz@scan@one@point\pgfutil@firstofone(#2)\relax},
        minimum width/.expanded=\the\pgf@x,
        minimum height/.expanded=\the\pgf@y}}
\makeatother

\title{\LARGE \bf
Autonomous Flight in Unknown GNSS-denied Environments for Disaster Examination
}

\author{Daniel Schleich, Marius Beul, Jan Quenzel, and Sven Behnke%
\thanks{This work has been supported by the German Federal Ministry of Education and Research (BMBF) in the project ``Kompetenzzentrum: Aufbau des Deutschen Rettungsrobotik-Zentrums (A-DRZ)''}%
\thanks{Institute for Computer Science VI, Autonomous Intelligent Systems, University of Bonn, Friedrich-Hirzebruch-Allee 8, 53115 Bonn, Germany,
		{\tt\small \{schleich, \ldots, behnke\}@ais.uni-bonn.de}%
}
}

\begin{document}

\maketitle
\thispagestyle{empty}
\pagestyle{empty}

\begin{tikzpicture}[remember picture,overlay]
\node[anchor=north west,align=left,font=\sffamily,xshift=0.5cm,yshift=-0.5cm] at (current page.north west) {%
  \footnotesize In: Proceedings of International Conference on Unmanned Aircraft Systems (ICUAS), Athens, Greece, June 2021.
};
\node[anchor=north east, align=right,font=\sffamily,xshift=-0.5cm,yshift=-0.5cm] at
(current page.north east) {%
  \footnotesize DOI: \href{https://doi.org/10.1109/ICUAS51884.2021.9476790}{10.1109/ICUAS51884.2021.9476790}
};
\end{tikzpicture}
\begin{abstract}
Micro aerial vehicles (MAVs) have high potential for information gathering tasks to support situation awareness in search and rescue scenarios.
Manually controlling MAVs in such scenarios requires experienced pilots and is error-prone, especially in stressful situations of real emergencies.
The conditions of disaster scenarios are also challenging for autonomous MAV systems.
The environment is usually not known in advance and GNSS might not always be available.

We present a system for autonomous MAV flights in unknown environments which does not rely on global positioning systems.
The method is evaluated in multiple search and rescue scenarios and allows for safe autonomous flights, even when transitioning between indoor and outdoor areas.
\end{abstract}

\section{Introduction}
\label{sec:Introduction}
Micro aerial vehicles (MAVs) are increasingly used in search and rescue scenarios for fast disaster examination and situation awareness.
They are mostly used to map outdoor environments, but MAVs also have potential for information extraction in indoor areas, especially in industrial environments:
Structurally damaged buildings or the presence of hazardous chemical substances, might prohibit exploration by humans, and the use of ground robots might be limited due to untraversable terrain.

In most applications, MAVs are remotely controlled by human operators.
However, when flying indoors or in the vicinity of obstacles, manually controlling a MAV is a challenging task, which requires an experienced pilot.
Since a direct line-of-sight between operator and MAV cannot always be maintained, communication latency and restricted awareness of the MAV's surroundings increase the difficulty even further.
Human errors or potential communication losses might result in crashes.

Autonomous flights help to reduce the strain on the operator, increase safety and make the applicability of MAVs feasible for less-experienced pilots.
However, many approaches for autonomous navigation either rely on GNSS-based localization or on pre-captured maps of the environment~\cite{beul2018fast}.
For real disaster scenarios in indoor environments, both are often not available.
In this work, we present our integrated system for safe, autonomous navigation in GNSS-denied, initially unknown environments.
Our framework includes
\begin{itemize}
 \item a method for precise LiDAR-based odometry in unknown environments,
 \item fast navigation planning considering MAV dynamics and optimizing for safety and flight time,
 \item and time-optimal trajectory generation and control.
\end{itemize}

Our system was evaluated in multiple scenarios, like outdoor and indoor flights and a simulated CBRNE-scenario.

\begin{figure}[t]
  \centering
  \includegraphics[width=0.8\linewidth]{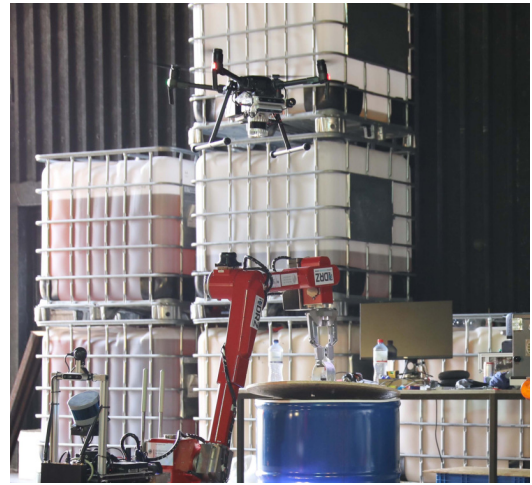}
  \caption{Our MAV guides the teleoperation of a ground robot in a simulated indoor CBRNE-scenario.}
  \label{fig:teaser}
\end{figure}
 
\section{Related Work}
\label{sec:Related_Work}
MAVs are commonly used in search and rescue scenarios for exploring large outdoor environments, \ie for forest fire monitoring~\cite{merino2012unmanned} or searching survivors during floods~\cite{ravichandran2019uav}.
In such scenarios, MAVs usually navigate a sequence of GPS-waypoints and are operated on sufficiently high altitudes such that no obstacle avoidance is necessary.
The use of MAVs for indoor fire detection has been studied by Pecho \etal\cite{pecho2019unmanned}.
However, autonomously navigating MAVs in such scenarios requires advanced localization and obstacle avoidance methods.

State estimation methods in GNSS-denied environments are mainly vision-based or LiDAR-based.
Mohta \etal\cite{mohta2018fast} employ visual odometry for autonomous flight in indoor areas.
During flights in warehouse-like environments, they encountered drift in the state estimation due to a low number of detectable visual features.
Mostafa \etal\cite{mostafa2018radar} address such challenges by using a combination of radar and visual odometry.

A different approach for autonomous flights in warehouses is proposed by Beul \etal\cite{beul2018fast}. 
It employs LiDAR-based localization but relies on a pre-captured environment map.
Spurny \etal\cite{spurny2021autonomous} use two different state estimation filters for outdoor and indoor environments.
The outdoor filter uses GNSS and magnetometer data while the indoor filter relies on simultaneous localization and mapping (SLAM) using a 2D LiDAR.
The combination of two different state estimators enables smooth transitions between outdoor and indoor environments.
A survey on different odometry methods for autonomous navigation in GNSS-denied environments has been compiled by Mohamed \etal\cite{mohamed2019survey}.
Our approach employs LiDAR-based odometry which does not require previous knowledge of the environment.

Planning collision-free trajectories in cluttered environments is commonly done in a two-stage approach:
A low-dimensional spatial path is planned first, and subsequently refined to a high-dimensional trajectory, \ie using B-Spline path planning~\cite{koyuncu2008probabilistic}, quadratic programming~\cite{richter2016polynomial} or gradient-based optimization~\cite{nieuwenhuisen2019search}.
The methods for indoor flights presented above also rely on low-dimensional path planning in 3D or 4D (position and yaw).

Refining position-only paths to high-dimensional trajectories only generates locally optimal solutions, though.
In contrast to these works, we directly incorporate the system dynamics into the planning to generate globally optimal trajectories with low execution times.
Dynamically feasible motion primitives are used to generate high-dimensional state lattices~\cite{liu2017search}, which are searched using A*.
In contrast to previous methods, we additionally apply local multiresolution~\cite{behnke2003local} to the state lattices:
The spatial resolution is decreased with increasing distance to the current MAV position.
Other approaches of using multiresolution with state lattices include the work of Gonz\'alez-Sieira \etal\cite{gonzalez2016adaptive}, where the resolution level is based on the complexity of the local environment.
Andersson \etal\cite{andersson2018receding} plan in high-dimensional space until a time threshold is reached, after which they continue planning in a lower-dimensional space.

An extension of the MAV controller used in our framework is presented in \cite{beul2020trajectory}.
For future work, we plan to integrate it into our navigation pipeline.
In this paper, we rely on fast trajectory replanning to react to newly perceived obstacles.
 
\section{System Setup}
\label{sec:system_setup}

\begin{figure}[t]
  \centering
  \resizebox{1.0\linewidth}{!}{%
\begin{tikzpicture}
[content_node/.append style={font=\sffamily,minimum size=1.5em,minimum width=6em,draw,align=center,rounded corners,scale=0.65},
label_node/.append style={font=\sffamily,scale=0.5},
group_node/.append style={font=\sffamily,dotted,align=center,rounded corners,inner sep=1em,thick},>={Stealth[inset=0pt,length=4pt,angle'=45]}]

\definecolor{red}{rgb}     {0.5,0.0,0.0}
\definecolor{green}{rgb}   {0.0,0.5,0.0}
\definecolor{blue}{rgb}    {0.0,0.0,0.5}
\definecolor{grey}{rgb}    {0.5,0.5,0.5}

\draw[thick, rounded corners, grey!20!white,fill] (-4.0,0.6) -- (4.75,0.6) -- (4.75,5.0) -- (-4.0,5.0) -- cycle;
\draw[thick, rounded corners, grey!20!white,fill] (-4.0,-1.75) -- (4.75,-1.75) -- (4.75,-0.25) -- (-4.0,-0.25) -- cycle;

\node(Operator)[content_node,fill=green!15!white] at (-3.0,4.5) {Operator};
\node(LIDAR)[content_node,fill=green!15!white] at (-3.0,3.5) {LiDAR};

\node(Mapping)[content_node,fill=blue!15!white] at (-.25,4.0) {Occupancy\\Mapping};
\node(Odometry)[content_node,fill=blue!15!white] at (-.25,3.0) {LiDAR\\Odometry};
\node(Filter)[content_node,fill=blue!15!white] at (-.25,1.5) {State\\Estimation};

\node(Planner)[content_node,fill=blue!15!white] at (2.9,4.5) {Trajectory\\Planner};
\node(Waypoint)[content_node,fill=blue!15!white] at (2.9,3.0) {Trajectory\\Tracking};
\node(Control)[content_node,fill=blue!15!white] at (2.9,1.5) {MAV\\Control};

\node(MAV)[content_node,fill=red!15!white] at (-.25,-1.0) {MAV};
\node(IMU)[content_node,fill=green!15!white] at (-3.0,-1.0) {IMU};

\draw[->, thick] (Operator) -- node[label_node,midway,below] {} node[label_node,midway,above] {Target Pose} (Operator -| 1.25, 4.5) -- ( 1.25,4.5 |- Planner.170) -- (Planner.170);

\draw[->,thick] (Mapping) -- node[label_node,midway,below] {\SI{10}{\hertz}} node[label_node,midway,above] {Obstacles} (Mapping -| 1.5,4.0) -- (1.5,-4.0 |- Planner.180) -- (Planner.180);

\draw[->,thick] (LIDAR) -- node[label_node,midway,below] {\SI{10}{\hertz}} node[label_node,midway,above] {Scan} (LIDAR -| -1.5,3.5) -- (-1.5,3.5 |- Mapping.180) -- (Mapping.180);
\draw[->, thick] (LIDAR) -- (LIDAR -| -1.5,3.5) -- (-1.5,3.5 |- Odometry.180) -- (Odometry.180);

\draw[->, thick] (Odometry) -- node[label_node,midway,left] {\SI{10}{\hertz}} node[label_node,midway,right,text width=3cm,yshift=0.0cm]{3D~MAV~Position} (Filter);

\draw[->,thick] (Filter) -- node[label_node,midway,below,text width=3cm, xshift=1.0cm] {3D~MAV~Pos. 3D~MAV~Vel. MAV~Roll MAV~Pitch MAV~Yaw} node[label_node,midway,above, xshift=0.5cm] {\SI{50}{\hertz}} (Filter -| 1.75,1.5) -- (1.75,1.5|- Planner.190) -- (Planner.190);
\draw[->,thick] (Filter) -- (Filter -| 1.75,1.5) -- (1.75,1.5|- Waypoint) -- (Waypoint);
\draw[->,thick] (Filter) --  (Control);

\draw[->, thick] (Planner) -- node[label_node,midway,left,yshift=0.0cm] {\SI{1}{\hertz}} node[label_node,midway,right,text width=3cm,yshift=0.0cm] {Trajectory} (Waypoint);

\draw[->, thick] (Waypoint) -- node[label_node,midway,left,yshift=0.0cm] {\SI{10}{\hertz}} node[label_node,midway,right,text width=3cm,yshift=0.0cm] {3D~Target~Position\\Target~Yaw} (Control);

\draw[->, thick] (Control) -- node[label_node,midway,left,yshift=0.0cm] {\SI{50}{\hertz}} node[label_node,midway,right,text width=1cm,yshift=0.0cm]{Roll Pitch Climb~rate Yaw~rate} (Control |- 2.9, -1.0 ) -- ( 2.9, -1.0 |- MAV.360 ) -- (MAV.360);

\draw[->, thick] (MAV)  -- node[label_node,midway,left,yshift=-0.5cm] {\SI{50}{\hertz}} node[label_node,midway,right,text width=2.8cm,yshift=-0.5cm] {3D~MAV~Accel. MAV~Roll MAV~Pitch MAV~Yaw} (Filter);

\draw[->, thick] (IMU) --  (IMU -| -1.75,-1.0) --  (-1.75,-1.0 |- MAV.180)  -- node[label_node,midway,left] {} node[label_node,midway,right] {} (MAV.180);

\node(ROS_Group_Label)[label_node,anchor=south west] at (-4.0,5.0) {\textbf{Onboard Computer}};
\node(MAV_Group_Label)[label_node,anchor=south west] at (-4.0,-0.25) {\textbf{DJI Matrice 210 v2}};

\end{tikzpicture}
}
  \caption{System overview.}
  \label{fig:system}
\end{figure}
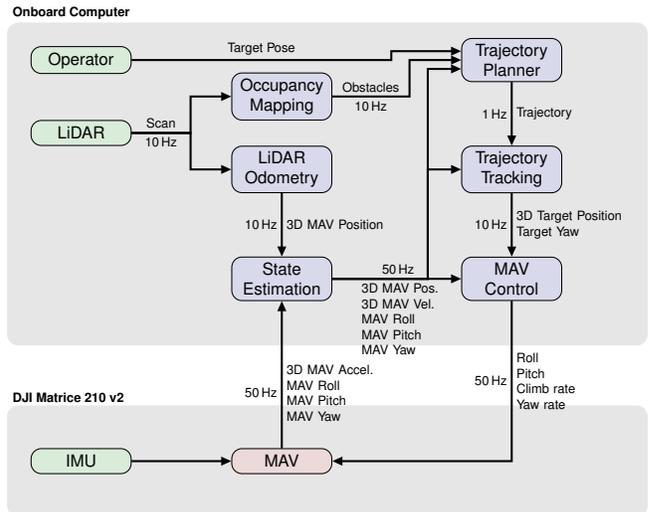

In the following sections, we describe the different components of our MAV system, an overview of which is depicted in \reffig{fig:system}.
First, we present our hardware setup in \refsec{sec:hardware}.
We continue by describing the environment perception modules, including occupancy mapping (\refsec{sec:mapping}), LiDAR-based odometry (\refsec{sec:odometry}), and a State Filter (\refsec{sec:state_filter}), which fuses IMU measurements and LiDAR odometry into an estimate of the high-dimensional MAV state.
Finally, we describe our navigation and control pipeline:
Safe, dynamically feasible trajectories are planned by searching a high-dimensional state lattice including velocities (\refsec{sec:planner}).
The next waypoint is sampled from the planned trajectory (\refsec{sec:waypoint_sampler}) and forwarded to our MAV controller (\refsec{sec:control}).

\subsection{Hardware Design}
\label{sec:hardware}

\begin{figure}[t]
  \centering
  \resizebox{1.0\linewidth}{!}{%
\begin{tikzpicture}
      [boxstyle/.style={font=\sffamily,black,fill=blue!20!white,fill opacity=0.8,text opacity=1,text=black,draw,ultra thick,align=center,rectangle callout}]
      \node[anchor=south east, inner sep = 0] (left_image) at (0,0) {\includegraphics[width=\linewidth]{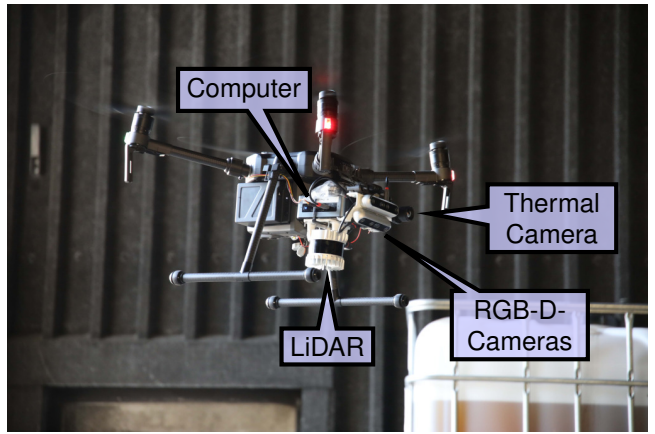}};
    \begin{scope}[shift=(left_image.south west),x={(left_image.south east)},y={(left_image.north west)}]
        \node[boxstyle,callout relative pointer={(0.08, -0.2)}] at (0.37,0.8) {Computer};
        \node[boxstyle,callout relative pointer={(0.0, 0.11)}] at (0.5,0.2) {LiDAR};
        \node[boxstyle,text width=1.5cm,callout relative pointer={(-0.09, 0.005)}] at (0.85,0.5) {Thermal Camera};
        \node[boxstyle,text width=1.6cm,callout relative pointer={(-0.13, 0.13)}] at (0.8,0.25) {RGB-D-Cameras};
    \end{scope}
\end{tikzpicture}
}
  \caption{Hardware design of our MAV.}
  \label{fig:hardware}
\end{figure}

\begin{figure*}
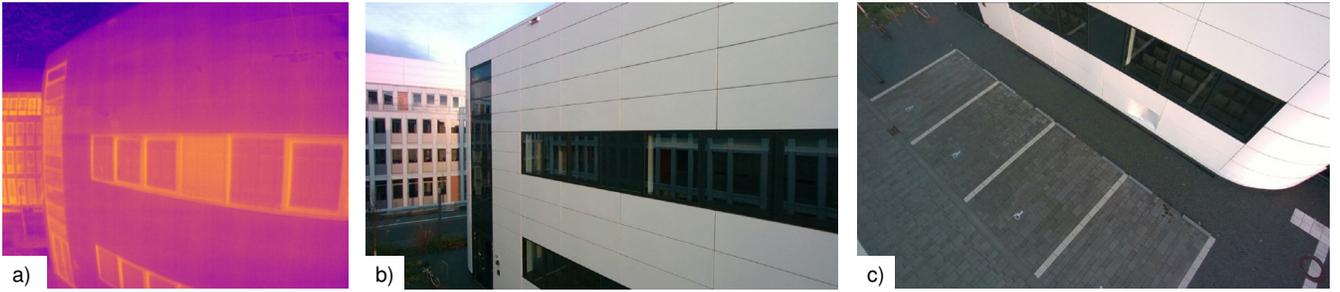

  \centering
  \resizebox{1.0\linewidth}{!}{%
  \begin{tikzpicture}
    \definecolor{red}{rgb}{0.7,0.0,0.0}
    \node[anchor=north west, inner sep=0] (image1) at (0, 0) {\includegraphics[trim=10mm 0mm 10mm 0mm,clip, width=0.227\textwidth]{./img/Onboard/poppelsdorf_thermal_jet.png}};
    \node[anchor=north west, inner sep=0] (image2) at (4.25, 0) {\includegraphics[trim=10mm 0mm 10mm 0mm,clip,width=0.31\textwidth]{./img/Onboard/poppelsdorf_front.png}};
    \node[anchor=north west, inner sep=0] (image3) at (10,0) {\includegraphics[trim=10mm 0mm 10mm 0mm,clip,width=0.31\textwidth]{./img/Onboard/poppelsdorf_down.png}};
    
    \node[label,scale=1.0, anchor=south west, rectangle, fill=white, align=center, font=\scriptsize\sffamily] (n_0) at (-0.01,-3.45) {a) };
    \node[label,scale=1.0, anchor=south west, rectangle, fill=white, align=center, font=\scriptsize\sffamily] (n_1) at (4.24,-3.45) {b)};
    \node[label,scale=1.0, anchor=south west, rectangle, fill=white, align=center, font=\scriptsize\sffamily] (n_2) at (9.99,-3.45) {c)};

  \end{tikzpicture}
}
  \caption{Onboard footage from Target 1 of the outdoor experiment described in \refsec{sec:poppelsdorf_eval}. a) Thermal camera. b) Forward-facing RGB-D camera. c) Downward-facing RGB-D camera.}
  \label{fig:onbard_footage}
\end{figure*}

We base our MAV (\reffig{fig:hardware}) on the commercially available DJI Matrice 210 v2 platform.
For onboard computing, we equip it with an Intel Bean Canyon NUC8i7BEH with Core i7-8559U processor and \SI{32}{\giga\byte} of RAM.
An Ouster OS-0 3D-LiDAR is used for odometry and obstacle detection.
To provide useful information in search and rescue missions, our MAV additionally features a FLIR AGX thermal camera for, \eg fire detection, and two Intel Realsense D455 RGB-D cameras which are mounted on top of each other to increase the vertical field of view.
An example of onboard footage from those sensors is shown in \reffig{fig:onbard_footage}.

\subsection{MARS LiDAR Odometry}
\label{sec:odometry}
\begin{figure*}
  \centering
  \resizebox{1.0\linewidth}{!}{\includegraphics{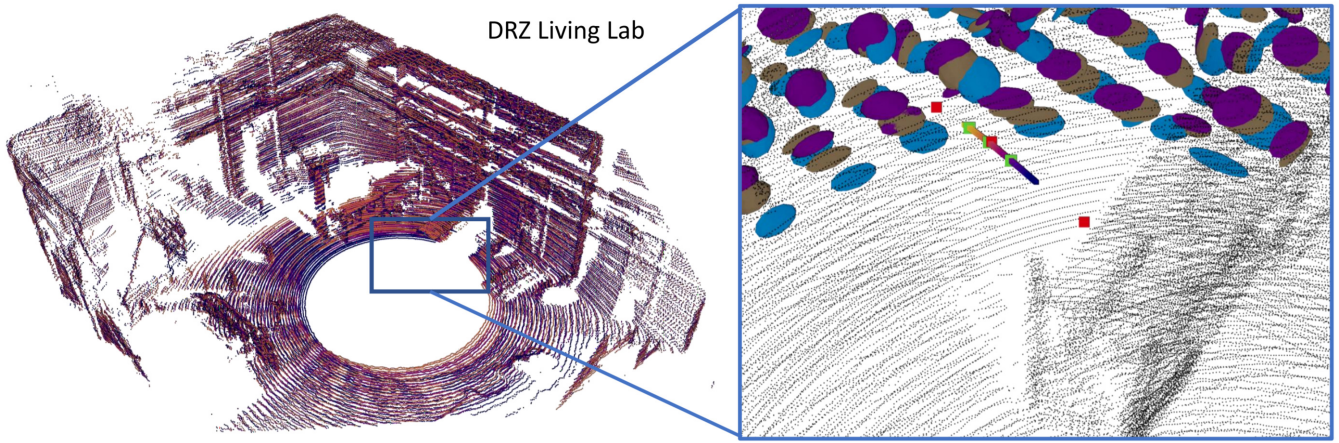}}
  \caption{Aligned point clouds of the sliding registration window described in \ref{sec:odometry}. Control points (red) define the spline (blue to yellow line) and interpolate the scan poses (green). The ellipsoids (surfel, colored per scan) describe the distribution of points in their vicinity.}
  \label{fig:odometry}
\end{figure*}

We used an early development version of our MARS LiDAR odometry~\cite{quenzel2021iros} during the experiments. We model surfaces within LiDAR scans with normal distributions derived from measured points on a uniform sparse voxel grid. This surface elements (surfel) map is created with multiple resolutions to increase detail close to the map origin. We half the side length and cell size per level for better representation of the sensor geometry. Our odometry uses a sliding registration window to simultaneously register multiple surfel maps against a local surfel map. \reffig{fig:odometry} shows an example from the DRZ Living Lab\footnote{\url{https://rettungsrobotik.de/living-lab/}}. A continuous-time Lie group B-Spline~\cite{sommer2020cvpr} describes the UAV trajectory within the sliding registration window. 
After a certain travelled distance, we add the last scan in a keyframe-based sliding window approach to the local surfel map and shift if necessary the local map to maintain its egocentric property.

\subsection{Environment Mapping}
\label{sec:mapping}
We perform a simple voxel-based occupancy mapping to enable high-level obstacle avoidance and trajectory planning. After registration, we transform the point cloud with the sensor pose into the local map frame and bin the points into voxels with \SI{25}{\centi\metre} side length. Then, we ray-trace with a 3D version of the Bresenham line-search~\cite{amanatides1987fast}, starting from the sensor pose towards the retained voxels. 
Every voxel in between start and end point that was not previously retained will be counted towards being empty, while every retained voxel will be counted towards being occupied. We improve the robustness against drift by removal of measurements that are older then $N=30$ scans, thus, reducing the time needed to measure the old obstacle location free.

\subsection{State Estimation}
\label{sec:state_filter}
The LiDAR-based odometry estimates the current MAV 3D position and orientation with a frequency of \SI{10}{\hertz}.
However, to enable closed-loop control, we need information about the MAV state at the control frequency of \SI{50}{\hertz}, including additional velocity estimates.
The MAV platform only provides GNSS-based velocity measurements, which are not available for indoor flights.
Thus, we use IMU acceleration measurements to estimate the velocity and correct these estimates using position information from LiDAR Odometry.

During all of our experiments, the IMU provided accurate 3D orientation data and the magnetometer was not affected by buildings or other metallic structures.
Since these measurements are provided at a much higher rate than our odometry data, we decided to only fuse IMU orientation measurements and omit LiDAR-based orientation data.

To fuse IMU and LiDAR-Odometry data, we utilize the implementation of an Extended Kalman Filter (EKF) from the robot\_localization library \cite{moore2014ekf}.
Input to the EKF are 3D positions from LiDAR-Odometry, and 3D orientations (which have to be transformed into the map frame first) and linear accelerations measured by the IMU.
The EKF outputs high-dimensional MAV states with \SI{50}{\hertz}.

\subsection{Trajectory Planning}
\label{sec:planner}
In disaster response scenarios, low mission execution times are essential, especially since MAV flight times are limited.
Thus, flight trajectories should not only be optimized with respect to safety but also for execution time.
This can be achieved by directly incorporating MAV dynamics into the planning.

\begin{figure}
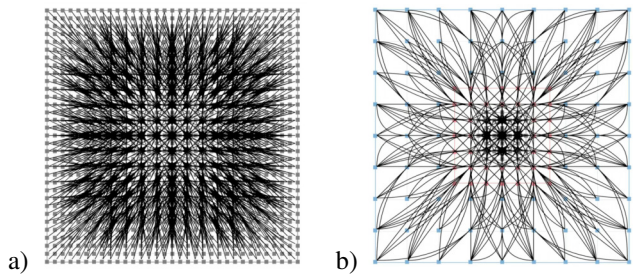

\begin{center}
\hspace*{-0.3em}a)\hspace{0.4em}\mbox{\includegraphics[width=0.2\textwidth]{img/uniform_lattice.png}}
  \hspace{0.4em}
b)\hspace{0.4em}\mbox{\includegraphics[width=0.2\textwidth]{img/mres_lattice.png}}
 \end{center}
 \caption{State lattice graphs. a) Uniform. b) Local multiresolution. The spatial position of nodes is fixed to the corners of a multiresolution grid with high resolution at the current MAV position, and coarser resolution for more distant areas. }
 \label{fig:state_lattices}
\end{figure}

We employ our trajectory planning method from \cite{schleich2021search}, which is based on the framework of Liu \etal \cite{liu2017search}.
The MAV state is modeled as a $6$-tuple $s=(\mathbf{p}, \mathbf{v}) \in \mathbb R^6$ consisting of a 3D position $\mathbf{p}$ and velocity $\mathbf{v}$.
A state lattice graph $\mathcal G$ is generated by unrolling motion primitives from the initial MAV state $s_0$.
Each primitive corresponds to applying a constant acceleration control $\mathbf{u}$ over a short time interval $\tau$, \ie can be expressed as a time-parameterized polynomial
\begin{equation*}
F_{\mathbf{u},s}(t) = \begin{pmatrix}\mathbf{p}+ t \mathbf{v} + \frac{t^2}{2} \mathbf{u}\\\mathbf{v} + t \mathbf{u}\end{pmatrix},\,\text{ for } t\in[0,\tau].
\label{eq:primitive}
\end{equation*}
Here, $s=(\mathbf{p}, \mathbf{v})$ denotes the initial state of the motion primitive and the corresponding action costs are defined as the weighted sum of control effort and primitive duration, \ie
\begin{equation*}
C(F_{\mathbf{u},s}) = ||\mathbf{u}||^2_2\tau + \rho\tau.
\label{eq:primitive_cost}
\end{equation*}
Trajectories are generated by applying A* search to the state lattice graph.

Searching high-dimensional state spaces is computationally expensive, but frequent replanning is necessary in unknown environments to react to newly perceived obstacles.
Thus, we extend the method of Liu \etal to achieve fast replanning times.
We restrict the state positions of the lattice graph $\mathcal G$ to the corners of a MAV-centered local multiresolution grid to significantly reduce the state space size.
An example is depicted in \reffig{fig:state_lattices}.
The MAV vicinity is represented at a high spatial resolution, while the resolution decreases with increasing distance to the MAV position.

Additionally, we propose a search heuristic that considers the resolution of the motion primitives by solving 1D sub-problems along the x-, y- and z-axis.
For each pair of signed distance to the goal position and start velocity, we precompute the costs of the optimal 1D trajectory without obstacle consideration.
During search, we combine the 1D costs into an estimate for the 3D costs: The flight time is calculated as the maximum over the individual axes, and the control costs are those of the sub-problem with highest execution time.
For further details on local multiresolutional state lattices and the proposed 1D heuristic, we refer to~\cite{schleich2021search}.

To increase the safety of the planned trajectories, we extend our previous work by adding additional obstacle costs to our planning framework.
The center points of the obstacles detected by the environment mapping module (\cf\refsec{sec:mapping}) are inserted into a k-d tree~\cite{bentley1975multidimensional} for fast distance queries.
During search, we sample positions on the motion primitives for collision checking.
If the distance $d(s,o)$ between a sample $s$ and the nearest obstacle $o$ is lower than a safety distance $d_\text{min}$, the position is assumed invalid.
For distances larger than an upper threshold $d_\text{max}$, no obstacle costs are added to the motion primitive.
Otherwise, we add costs $c$ that linearly decrease with increasing distance to the nearest obstacle, \ie we set $c=\frac{ d_\text{max} - d(s,o) }{ d_\text{max} - d_\text{min} }$.

Due to trajectory tracking errors, the MAV might enter the safety area around obstacles.
Replanning would fail in such cases, since the initial MAV position is invalid.
Therefore, we allow invalid start positions during collision checking, but require that the obstacle distance between two adjacent position samples does not decrease.
By adding high obstacle costs to invalid positions, we ensure that the MAV immediately returns to a safe distance on the shortest possible trajectory.

To react to newly perceived obstacles, replanning is triggered at \SI{1}{\hertz}.
This frequency was empirically determined from maximum replanning times.
As described in \cite{schleich2021search}, replanning takes less than \SI{1}{\second} in most of the cases, even for large environments.
For smaller environments, a higher replanning frequency is possible.

As described in \refsec{sec:waypoint_sampler}, a waypoint is sampled from the current trajectory and forwarded to the MAV controller.
The start state $s_\text{replan}$ for replanning is selected from the trajectory at the time $t_\text{plan}$ ahead of the current waypoint.
Here, $t_\text{plan}$ corresponds to the estimated replanning duration.
The trajectory from the current MAV state up to $s_\text{replan}$ is not updated.

\subsection{Trajectory Tracking}
\label{sec:waypoint_sampler}
Our planning framework generates second-order trajectories, which need to be further refined for execution.
Thus, we sample waypoints on the planned trajectory with a fixed timestep of \SI{0.1}{\second}.
The first waypoint $p_0$ is selected at a fixed time interval $t_0$ after the trajectory start and is forwarded as a goal pose to the MAV controller.
Afterwards, we keep publishing waypoints at a frequency of \SI{10}{\hertz} while moving them accordingly along the trajectory.
If the distance between the MAV and the current waypoint $p_i$ exceeds a threshold $d_\text{tracking}$, the MAV was not able to track the trajectory.
In this situation, we distinguish between two cases:
\begin{enumerate}
 \item The MAV tracks the trajectory spatially but at a lower velocity.
 \item The MAV does not track the trajectory spatially.
\end{enumerate}
In the first case, we wait until the MAV reaches a position sufficiently close to $p_i$ before continuing to publish the next waypoints.
In the second case, the distance between the MAV and the planned flight path exceeds a threshold $d_\text{replan}$.
Thus, we abort trajectory tracking and restart by planning a new trajectory, starting from the current MAV position.

Note that we do not aim at precisely executing the planned trajectory.
Due to sampling the waypoints ahead in time, the MAV changes flight direction slightly before the trajectory does.
These tracking errors do not affect the MAV safety since all planned trajectories keep sufficiently large safety margins to obstacles.

\subsection{MAV Control} 
\label{sec:control}
Reliably approaching the sampled waypoint close to structures despite external disturbances is a challenging task. For precise MAV control, we employ our method presented in \cite{beul2016icuas} with the extensions from \cite{beul2017icuas}. For brevity, in this section, we cover only the most essential aspects of the algorithm.

Our technique models the MAV as a multidimensional triple integrator with nonlinear state boundaries and jerk as system input. Based on this model, we analytically generate third-order time-optimal trajectories that satisfy asymmetrical input ($j_{min} \leq j \leq j_{max}$) and state constraints ($a_{min} \leq a \leq a_{max}$, $v_{min} \leq v \leq v_{max}$). Trajectories are computed from the current state $(p,v,a)_{MAV}^\intercal$ to the sampled waypoint state $(p,0,0)_{wayp}^\intercal$ with zero target velocity and acceleration. We temporally synchronize the x-, y- and z-axis to arrive at the target state at the same time.

Trajectories are replanned with \SI{50}{\hertz} and are directly executed by the MAV serving as model predictive controller (MPC). As stated above, our MPC generates jerk commands, but the low-level flight controller expects pitch resp. roll commands. Therefore, we assume pitch and roll to relate to $\theta = \atantwo(a_x,g)$ and $\phi = \atantwo(a_y,g)$. Thus, we send smooth pitch $\theta$ and roll $\phi$ commands for horizontal movement and smooth climb rates $v_{z}$ instead.

MAVs are inherently unstable (triple integrator model); thus, closed-loop control is necessary to stabilize the system at all times. Since many demanding tasks are running onboard the MAV's computer, we assign real-time priority for the MPC to ensure execution of this vital task.

\begin{figure}[t]
  \centering
  \includegraphics[trim=10mm 0mm 10mm 0mm, width=0.8\linewidth]{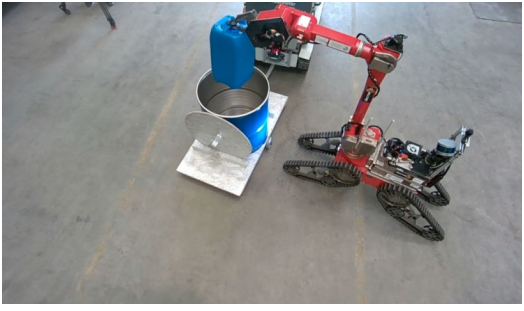}
  \caption{Onboard footage during a CBRNE-scenario. By providing a top-down view of the workspace, our MAV facilitates the tele-operation of a ground robot, which has to secure hazardous substances.}
  \label{fig:CBRNE_onboard_view}
\end{figure}
 
\section{Evaluation}
\label{sec:Evaluation}
We employed our MAV in two simulated CBRNE-scenarios in cooperation with firefighters and multiple tele-operated ground robots.
Our MAV was used to generate an overview of the situation, while the ground robots were used to secure hazardous substances.
In both scenarios, our MAV proved useful in assisting the operators of the ground robots by providing additional top-down views of the workspace.
Figure \ref{fig:teaser} shows a scene during the first CBRNE-scenario and onboard footage of the second scenario is provided in \reffig{fig:CBRNE_onboard_view}.

In the following sections, we evaluate the components of our framework.
Here, we concentrate on the interaction between the different components and on the overall performance of our system.
A more detailed evaluation of the individual components, including computation times and benchmarks against other methods, can be found in the corresponding papers~\cite{quenzel2021iros, schleich2021search, beul2017icuas}.

\begin{figure}[t]
  \centering
  \resizebox{1.0\linewidth}{!}{%
\begin{tikzpicture}
      [boxstyle/.style={font=\sffamily,black,fill=blue!20!white,fill opacity=0.4,text opacity=1,text=black,draw,ultra thick,align=center,rectangle callout}]
      \node[anchor=south east, inner sep = 0] (left_image) at (0,0) {\includegraphics[width=\linewidth]{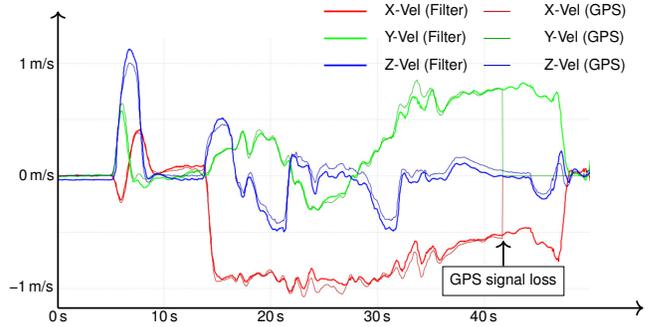}};
      
      \definecolor {x_color}{RGB}{255, 0, 0};
      \definecolor {y_color}{RGB}{0, 255, 0};
      \definecolor {z_color}{RGB}{0, 0, 255};
      \definecolor {x_color2}{RGB}{170, 0, 0};
      \definecolor {y_color2}{RGB}{0, 170, 0};
      \definecolor {z_color2}{RGB}{0, 0, 170};
      
    \begin{scope}[shift=(left_image.south west),x={(left_image.south east)},y={(left_image.north west)}]
    
    \node[label,scale=1.0, align=center, font=\scriptsize\sffamily] (t0) at (0,-0.03) {\SI{0}{\second}};
    \node[label,scale=1.0, align=center, font=\scriptsize\sffamily] (t1) at (0.2,-0.03) {\SI{10}{\second}};
    \node[label,scale=1.0, align=center, font=\scriptsize\sffamily] (t2) at (0.4,-0.03) {\SI{20}{\second}};
    \node[label,scale=1.0, align=center, font=\scriptsize\sffamily] (t3) at (0.6,-0.03) {\SI{30}{\second}};
    \node[label,scale=1.0, align=center, font=\scriptsize\sffamily] (t4) at (0.8,-0.03) {\SI{40}{\second}};
    
    \node[label,scale=1.0, align=center, font=\scriptsize\sffamily] (v0) at (-0.04, 0.495) {\SI{0}{\meter\per\second}};
    \node[label,scale=1.0, align=center, font=\scriptsize\sffamily] (v1) at (-0.05, 0.08) {\SI{-1}{\meter\per\second}};
    \node[label,scale=1.0, align=center, font=\scriptsize\sffamily] (v2) at (-0.04, 0.907) {\SI{1}{\meter\per\second}};
    
    \node[label,scale=1.0, align=center, rectangle, draw, fill=white, font=\scriptsize\sffamily] (enter) at (0.837,0.1) {GPS signal loss};
    \draw[thick,->] (enter) -> (0.837,0.25);

    \draw[thick, ->] (0,0) -- (1.1,0);
    \draw[thick, ->] (0,0) -- (0,1.1);

    \draw[thick, color=x_color] (0.5,1.1) -- (0.58,1.1);
    \node[label,scale=1.0, anchor=west, font=\scriptsize\sffamily] at (0.6, 1.1) {X-Vel (Filter)};
    \draw[color=x_color2] (0.8,1.1) -- (0.85,1.1);
    \node[label,scale=1.0, anchor=west, font=\scriptsize\sffamily] at (0.9, 1.1) {X-Vel (GPS)};
    
    \draw[thick, color=y_color] (0.5,1.0) -- (0.58,1.0);
    \node[label,scale=1.0, anchor=west, font=\scriptsize\sffamily] at (0.6, 1.0) {Y-Vel (Filter)};
    \draw[color=y_color2] (0.8,1.0) -- (0.85,1.0);
    \node[label,scale=1.0, anchor=west, font=\scriptsize\sffamily] at (0.9, 1.0) {Y-Vel (GPS)};
    
    \draw[thick, color=z_color] (0.5,0.9) -- (0.58,0.9);
    \node[label,scale=1.0, anchor=west, font=\scriptsize\sffamily] at (0.6, 0.9) {Z-Vel (Filter)};
    \draw[color=z_color2] (0.8,0.9) -- (0.85,0.9);
    \node[label,scale=1.0, anchor=west, font=\scriptsize\sffamily] at (0.9, 0.9) {Z-Vel (GPS)};

    \end{scope}
\end{tikzpicture}
}
  \caption{Comparison of estimated velocities of our state filter and ground truth GPS velocities during a part of the flight described in \refsec{sec:lbh_eval}. Note that estimation is robust to GPS loss.}
  \label{fig:filter_output}
\end{figure}

\begin{figure}[t]
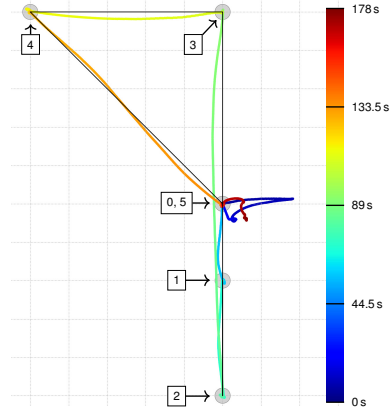

  \centering
  \resizebox{0.6\linewidth}{!}{%
\begin{tikzpicture}
      [boxstyle/.style={font=\sffamily,black,fill=blue!20!white,fill opacity=0.4,text opacity=1,text=black,draw,ultra thick,align=center,rectangle callout}]
      \node[anchor=south east, inner sep = 0] (left_image) at (0,0) {\includegraphics[width=0.8\linewidth]{./img/Control_test/without_gps.png}};
    \begin{scope}[shift=(left_image.south west),x={(left_image.south east)},y={(left_image.north west)}]
      \node[anchor=south west, inner sep = 0] (color_map) at (0.95,0.01) {\includegraphics[trim= 12px 0px 12px 0px, clip, width=0.042\linewidth]{./img/color_map.png}};
        \node[label,scale=1.0, align=center, rectangle, draw, fill=white, font=\scriptsize\sffamily] (n_0) at (0.5,0.5) {0, 5};
        \draw[thick,->] (n_0) -> (0.6,0.5);
        
        \node[label,scale=1.0, align=center, rectangle, draw, fill=white, font=\scriptsize\sffamily] (n_1) at (0.5,0.31) {1};
        \draw[thick,->] (n_1) -> (0.6,0.31);
        
        \node[label,scale=1.0, align=center, rectangle, draw, fill=white, font=\scriptsize\sffamily] (n_2) at (0.5,0.025) {2};
        \draw[thick,->] (n_2) -> (0.6,0.025);
        
        \node[label,scale=1.0, align=center, rectangle, draw, fill=white, font=\scriptsize\sffamily] (n_3) at (0.55,0.89) {3};
        \draw[thick,->] (n_3) -> (0.62,0.95);
        
        \node[label,scale=1.0, align=center, rectangle, draw, fill=white, font=\scriptsize\sffamily] (n_4) at (0.06,0.89) {4};
        \draw[thick,->] (n_4) -> (0.06,0.94);

    \end{scope}
    \begin{scope}[shift=(color_map.south west),x={(color_map.south east)},y={(color_map.north west)}]
        \node[label,scale=1.0, align=center, font=\scriptsize\sffamily, anchor=west] (end) at (1.1,1) {\SI{178}{\second}};
        \node[label,scale=1.0, align=center, font=\scriptsize\sffamily, anchor=west] (three_quater) at (1.1,0.75) {\SI{133.5}{\second}};
        \node[label,scale=1.0, align=center, font=\scriptsize\sffamily, anchor=west] (half) at (1.1,0.5) {\SI{89}{\second}};
        \node[label,scale=1.0, align=center, font=\scriptsize\sffamily, anchor=west] (quater) at (1.1,0.25) {\SI{44.5}{\second}};
        \node[label,scale=1.0, align=center, font=\scriptsize\sffamily, anchor=west] (start) at (1.1,0.) {\SI{0}{\second}};
        
        \draw[thick] (0,0) -- (1,0);
        \draw[thick] (0,0.25) -- (1,0.25);
        \draw[thick] (0,0.5) -- (1,0.5);
        \draw[thick] (0,0.75) -- (1,0.75);
        \draw[thick] (0,1) -- (1,1);

    \end{scope}
\end{tikzpicture}
}
  \caption{Top-down view of the flight path during our control experiment. Target positions are marked with gray circles and the optimal trajectory is depicted in black. The background grid has a resolution of \SI{1}{\meter}.}
  \label{fig:control_test}
\end{figure}

In all experiments, environment maps for collision avoidance and localization are generated online during flight.
No prior knowledge about the environment is necessary.
Since our MAV features a 3D-LiDAR with a vertical field of view of \SI{90}{\degree}, many obstacles are perceived while the MAV is still on the ground.
Thus, initial environment maps are already available before takeoff.
Target poses are manually defined by placing a marker in those maps.

\begin{figure*}[t]
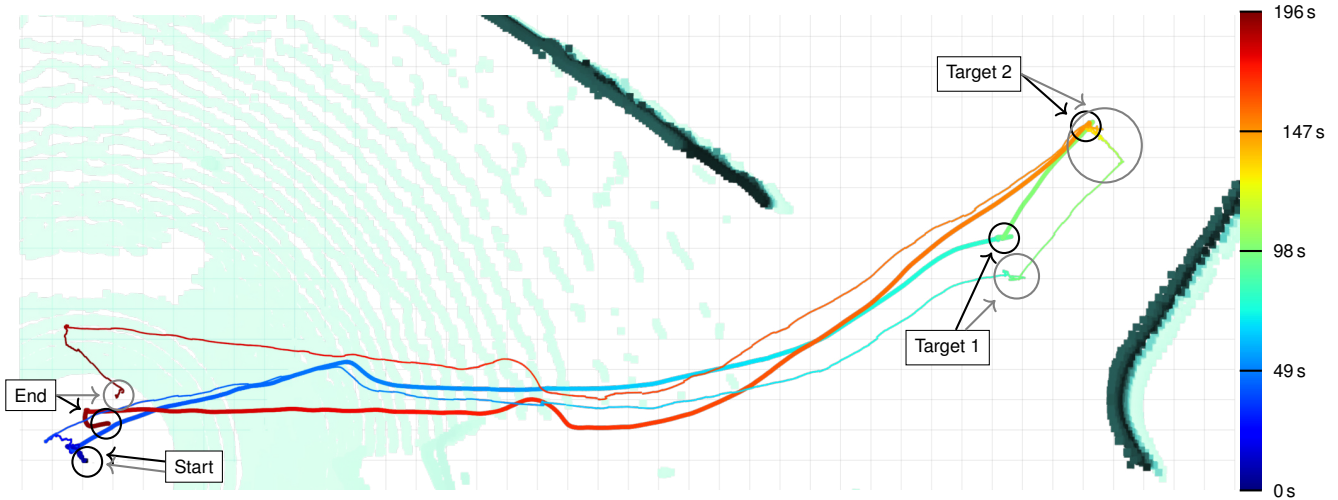

  \centering
  \resizebox{1.0\linewidth}{!}{%
\begin{tikzpicture}
      [boxstyle/.style={font=\sffamily,black,fill=blue!20!white,fill opacity=0.4,text opacity=1,text=black,draw,ultra thick,align=center,rectangle callout}]
      
      \definecolor{gps}{rgb}{0.5,0.5,0.5}
      \node[anchor=south east, inner sep = 0] (left_image) at (0,0) {\includegraphics[trim= 0px 0px 20px 0px, clip, width=0.95\linewidth]{./img/Poppelsdorf/gps_lidar_overlay.png}};
    \begin{scope}[shift=(left_image.south west),x={(left_image.south east)},y={(left_image.north west)}]
      \node[anchor=south west, inner sep = 0] (color_map) at (0.9825,0) {\includegraphics[trim= 12px 0px 12px 0px, clip, width=0.016\linewidth]{./img/color_map.png}};
        \node[label,scale=1.0, align=center, rectangle, draw, fill=white, font=\scriptsize\sffamily] (n_0) at (0.75,0.3) {Target 1};
        \draw[thick,->] (n_0) -> (0.785,0.5);
        \draw[thick,->,gps] (n_0) -> (0.788,0.4);
        
        \draw[black, thick] (0.795,0.53) circle (0.2cm);
        \draw[gps, thick] (0.805,0.45) circle (0.3cm);
        
        \node[label,scale=1.0, align=center, rectangle, draw, fill=white, font=\scriptsize\sffamily] (n_1) at (0.775,0.875) {Target 2};
        \draw[thick,->] (n_1.360) -> (0.845,0.79);
        \draw[thick,->, gps] (n_1.360) -> (0.86,0.81);
        
        \draw[black, thick] (0.86,0.765) circle (0.2cm);
        \draw[gps, thick] (0.875,0.725) circle (0.5cm);
        
        \node[label,scale=1.0, align=center, rectangle, draw, fill=white, font=\scriptsize\sffamily] (n_2) at (0.15,0.05) {Start};
        \draw[thick,->] (n_2.170) -> (0.08,0.075);
        \draw[thick,->, gps] (n_2.190) -> (0.08,0.055);

        \draw[black, thick] (0.065,0.06) circle (0.2cm);

        \node[label,scale=1.0, align=center, rectangle, draw, fill=white, font=\scriptsize\sffamily] (n_3) at (0.02,0.2) {End};
        \draw[thick,->] (n_3.360) -> (0.06,0.17);
        \draw[thick,->,gps] (n_3.360) -> (0.075,0.2);
        
        \draw[black, thick] (0.08,0.14) circle (0.2cm);
        \draw[gps, thick] (0.09,0.2) circle (0.2cm);

    \end{scope}
    \begin{scope}[shift=(color_map.south west),x={(color_map.south east)},y={(color_map.north west)}]
        \node[label,scale=1.0, align=center, font=\scriptsize\sffamily, anchor=west] (end) at (1.1,1) {\SI{196}{\second}};
        \node[label,scale=1.0, align=center, font=\scriptsize\sffamily, anchor=west] (three_quater) at (1.5,0.75) {\SI{147}{\second}};
        \node[label,scale=1.0, align=center, font=\scriptsize\sffamily, anchor=west] (half) at (1.1,0.5) {\SI{98}{\second}};
        \node[label,scale=1.0, align=center, font=\scriptsize\sffamily, anchor=west] (quater) at (1.1,0.25) {\SI{49}{\second}};
        \node[label,scale=1.0, align=center, font=\scriptsize\sffamily, anchor=west] (start) at (1.1,0.) {\SI{0}{\second}};
        
        \draw[thick] (0,0) -- (1,0);
        \draw[thick] (0,0.25) -- (1,0.25);
        \draw[thick] (0,0.5) -- (1,0.5);
        \draw[thick] (0,0.75) -- (1,0.75);
        \draw[thick] (0,1) -- (1,1);

    \end{scope}
\end{tikzpicture}
}
  \caption{Comparison of our LiDAR-based odometry (thick) and GPS (thin) during execution of the outdoor mission (top-down view). The parts of the trajectory corresponding to start, end and the two target poses are encircled (black for LiDAR-based odometry and gray for GPS). The background depicts the initial obstacle map. Darker color indicates a larger height value. The background grid has a resolution of \SI{1}{\meter}. }
  \label{fig:localization_poppelsdorf}
\end{figure*}

We analyze data from multiple autonomous flights captured during simulated search and rescue scenarios.
Our system works in outdoor and indoor environments, including transitions between both.
Since we fly close to structures in initially unknown environments, we chose to restrict the maximum flight velocity to \SI{1}{\meter\per\second}.

\subsection{State Estimation and Control}
\label{sec:control_eval}

\begin{figure}[t]
  \centering
  \includegraphics[width=\linewidth]{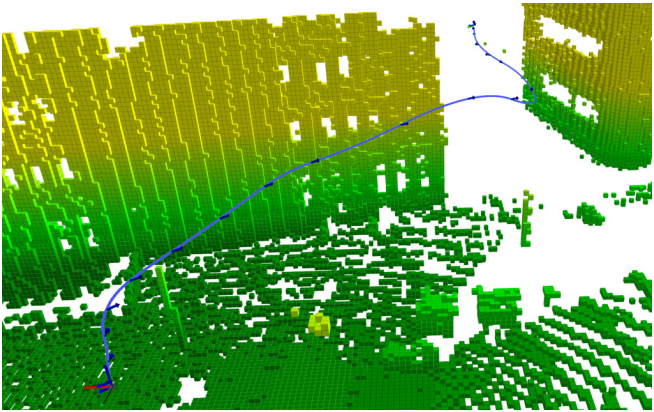}
  \caption{A planned trajectory in the outdoor scenario. Blue arrows depict the flight direction. The temporal distance between two consecutive arrows is \SI{0.5}{\second}.}
  \label{fig:trajectory_poppelsdorf}
\end{figure}

Reliably controlling the MAV close to structures requires precise state information.
Thus, we evaluate whether our state estimation (\cf\refsec{sec:state_filter}) is sufficiently accurate to be used with the MAV controller (\cf\refsec{sec:control}) for precise navigation.
As described above, our state filter infers velocities from fused position and acceleration measurements.
In a first experiment, we compare the estimated velocities against velocities measured by the onboard DJI GPS, which we consider as ground truth.
The corresponding data was captured during an autonomous flight from the outside into an industrial building, which will be detailed in \refsec{sec:lbh_eval}.
Figure \ref{fig:filter_output} shows that the estimated velocities align with the GPS-based ground truth until we enter the building.
While our state filter continues to generate velocity estimates, no GPS-data is available inside the building.
Note that the DJI interface still provides z-velocities since those are not only inferred from GPS but also from barometric data and a downward facing ultrasonic sensor.

In a second experiment, we evaluate MAV control based on the state estimation.
We manually took off and then switched to autonomous control.
We commanded the MAV to five different target positions, as shown in \reffig{fig:control_test}.
The MAV precisely reached all of the targets without overshooting.
Furthermore, the position at each waypoint was stably maintained for several seconds.

\subsection{Outdoor Flight}
\label{sec:poppelsdorf_eval}

In this experiment, we apply our system in an outdoor scenario where the MAV has to observe a facade from two different manually defined poses.
The MAV starts at the ground and has to move into a passage between two buildings.
An example for a planned trajectory is depicted in \reffig{fig:trajectory_poppelsdorf}.
This scenario is challenging for autonomous navigation since the GPS signal quality is significantly reduced in the narrow passage between the two buildings.
However, our approach successfully guides the MAV on a safe trajectory towards both target poses and back again due to our robust LiDAR-based odometry.
We compare the localization with our approach against GPS in \reffig{fig:localization_poppelsdorf}.
Especially when hovering at Target 2, which is located at the center of the narrow passage, GPS shows significant drift while our odometry correctly estimates that the MAV is not moving.

\subsection{Combined Outdoor and Indoor Flight} 
\label{sec:lbh_eval}

\begin{figure*}[t]
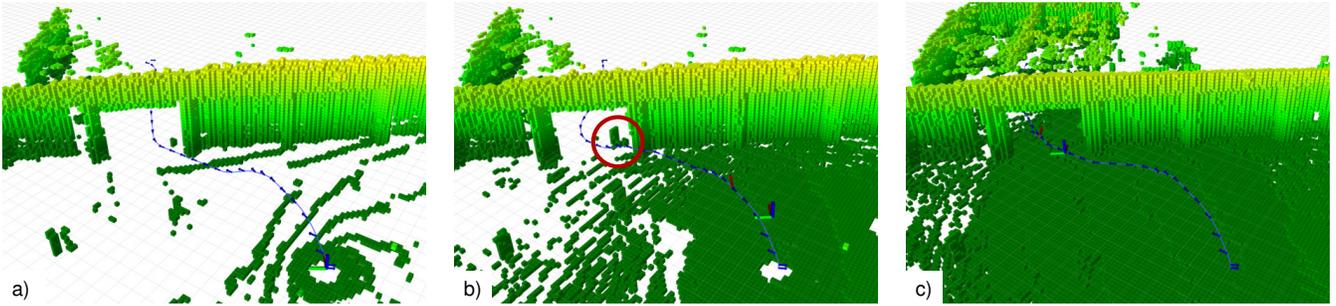

  \centering
  \resizebox{1.0\linewidth}{!}{%
  \begin{tikzpicture}
    \definecolor{red}{rgb}{0.7,0.0,0.0}
    \node[anchor=north west, inner sep=0] (image1) at (0, 0) {\includegraphics[trim=10mm 0mm 10mm 0mm,clip, width=0.29\textwidth]{./img/LBH/Path1.png}};
    \node[anchor=north west, inner sep=0] (image2) at (5.5, 0) {\includegraphics[trim=10mm 0mm 10mm 0mm,clip,width=0.29\textwidth]{./img/LBH/Path3.png}};
    \node[anchor=north west, inner sep=0] (image3) at (11,0) {\includegraphics[trim=10mm 0mm 10mm 0mm,clip,width=0.29\textwidth]{./img/LBH/Path5.png}};
    
    \node[label,scale=1.0, anchor=south west, rectangle, fill=white, align=center, font=\scriptsize\sffamily] (n_0) at (0,-3.75) {a) };
    \node[label,scale=1.0, anchor=south west, rectangle, fill=white, align=center, font=\scriptsize\sffamily] (n_1) at (5.5,-3.75) {b)};
    \node[label,scale=1.0, anchor=south west, rectangle, fill=white, align=center, font=\scriptsize\sffamily] (n_2) at (11,-3.75) {c)};

    \draw[line width=0.5mm, red] (7.5, -1.7) circle (0.3cm and 0.3cm);

  \end{tikzpicture}
}
  \caption{Replanning during the indoor experiment. A red arrow corresponds to the currently sampled waypoint for the MAV controller. Blue arrows depict the flight direction. The temporal distance between two consecutive blue arrows is \SI{0.5}{\second}. Note how the trajectory is updated to react to a dynamic obstacle (red circle).}
  \label{fig:flight_lbh}
\end{figure*}

\begin{figure}[t]
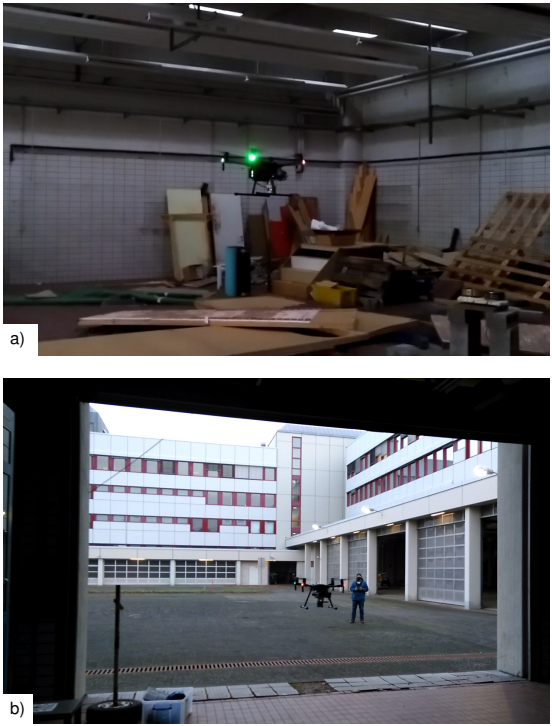

\centering
  \resizebox{0.85\linewidth}{!}{%
  \begin{tikzpicture}
    \definecolor{red}{rgb}{0.7,0.0,0.0}
    \node[anchor=north west, inner sep=0] (image1) at (0, 0) {\includegraphics[width=0.9\textwidth]{./img/LBH/inside.png}};
    \node[anchor=north west, inner sep=0] (image2) at (0.0, -11) {\includegraphics[width=0.9\textwidth]{./img/LBH/exit.png}};
    
    \node[label,scale=2.0, anchor=south west, rectangle, fill=white, align=center, font=\scriptsize\sffamily] (n_0) at (-0.05,-10.35) {a) };
    \node[label,scale=2.0, anchor=south west, rectangle, fill=white, align=center, font=\scriptsize\sffamily] (n_1) at (-0.05,-21.2) {b)};

  \end{tikzpicture} 
}
 \caption{Our MAV during autonomous flight. a) Reaching the observation pose inside an industrial building. b)Autonomously exiting the building through the gate while being supervised by a safety pilot. }
 \label{fig:mav_img_lbh}
\end{figure}

\begin{figure}[t]
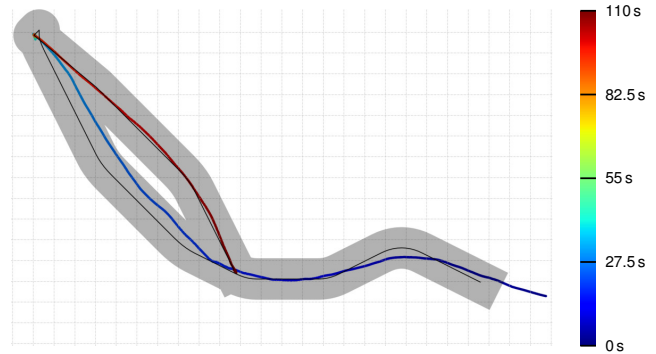

  \centering
  \resizebox{1.0\linewidth}{!}{%
\begin{tikzpicture}
      [boxstyle/.style={font=\sffamily,black,fill=blue!20!white,fill opacity=0.4,text opacity=1,text=black,draw,ultra thick,align=center,rectangle callout}]
      \node[anchor=south east, inner sep = 0] (left_image) at (0,0) {\includegraphics[width=\linewidth]{./img/LBH/LBH_tracking_xy2.png}};
    \begin{scope}[shift=(left_image.south west),x={(left_image.south east)},y={(left_image.north west)}]
      \node[anchor=south west, inner sep = 0] (color_map) at (1.05,0) {\includegraphics[trim=12px 0px 12px 0px, clip, width=0.0273\linewidth]{./img/color_map.png}};

    \end{scope}
    \begin{scope}[shift=(color_map.south west),x={(color_map.south east)},y={(color_map.north west)}]
        \node[label,scale=1.0, align=center, font=\scriptsize\sffamily, anchor=west] (end) at (1.1,1) {\SI{110}{\second}};
        \node[label,scale=1.0, align=center, font=\scriptsize\sffamily, anchor=west] (three_quater) at (1.1,0.75) {\SI{82.5}{\second}};
        \node[label,scale=1.0, align=center, font=\scriptsize\sffamily, anchor=west] (half) at (1.1,0.5) {\SI{55}{\second}};
        \node[label,scale=1.0, align=center, font=\scriptsize\sffamily, anchor=west] (quater) at (1.1,0.25) {\SI{27.5}{\second}};
        \node[label,scale=1.0, align=center, font=\scriptsize\sffamily, anchor=west] (start) at (1.1,0.) {\SI{0}{\second}};
        
        \draw[thick] (0,0) -- (1,0);
        \draw[thick] (0,0.25) -- (1,0.25);
        \draw[thick] (0,0.5) -- (1,0.5);
        \draw[thick] (0,0.75) -- (1,0.75);
        \draw[thick] (0,1) -- (1,1);

    \end{scope}
\end{tikzpicture} 
}
  \caption{Trajectory of the combined outdoor/ indoor scenario. The actual flight trajectory is colorized with respect to flight time. The black line connects the waypoints sampled from the planned trajectory. Since the first waypoint is sampled $t_0=$\SI{1}{\second} ahead of the initial position, there is some offset to the start of the flight trajectory. The gray area marks a tunnel with radius \SI{1}{\meter} around the planned trajectory. The background grid has a resolution of \SI{1}{\meter}.}
  \label{fig:tracking_lbh}
\end{figure}

In a last experiment, we apply our system in a scenario where the MAV has to explore the inside of an industrial building.
The MAV starts outside and has to enter through a gate to reach a manually defined observation pose.
There, it rotates to generate an overview of the environment and leaves the building again.
Example images of the flight are shown in \reffig{fig:mav_img_lbh}.
The full mission from takeoff until we left the building again took \SI{105}{\second}.
Since we plan shorter trajectories and have to find a path through a narrow doorway, we disable local multiresolution planning and use uniform state lattices instead, which is more suitable in such scenarios.

Additionally, we evaluate how our framework reacts to newly perceived obstacles.
The initially planned trajectory is shown in \reffig{fig:flight_lbh}\,a.
After takeoff, a person moves into the doorway.
The new obstacle is added to the occupancy map (red circle in \reffig{fig:flight_lbh}\,b).
The trajectory is updated accordingly to avoid a collision.
In \reffig{fig:flight_lbh}\,c, the person has moved away and the corresponding cells in the obstacle map are cleared again.
Since the MAV has already committed to a point in the doorway (red arrow), only parts of the trajectory inside the building are updated.

Finally, we evaluate the accuracy of our trajectory tracking method.
Figure \ref{fig:tracking_lbh} compares the actual flight trajectory against the planned trajectory.
As expected, the MAV does not perfectly track the planned trajectory in curves, due to sampling the waypoints ahead in time.
However, the tracking error is always below \SI{1}{\meter} and is thus covered by the larger security margin to obstacles.
The MAV accurately tracks the trajectory on straight segments.

Note that no motion capture system was available to evaluate the accuracy of our localization inside the industrial building of this flight experiment.
However, Quenzel and Behnke~\cite{quenzel2021iros} present a corresponding experiment of an indoor flight inside the DRZ Living Lab (\cf\reffig{fig:odometry}), which is equipped with a motion capture system.
 
\section{Conclusion}
\label{sec:Conclusion}
In this paper, we provided detailed insight into our navigation framework for safe, autonomous flight in GNSS-denied environments.
We showed the viability of our approach by employing our MAV system in multiple search and rescue scenarios.
Our proposed method for LiDAR-based odometry allows safe indoor flights.
Furthermore, it increases navigation accuracy when flying close to structures in outdoor environments where GPS signal quality is reduced.
Fast replanning and precise control enable safe flights in dynamic environments.
No previous knowledge of the environment is necessary since all maps are built online during flight.
Thus, our system addresses many challenges encountered in real disaster-response scenarios.
 
\bibliographystyle{IEEEtranBST/IEEEtran}
\bibliography{literature}

\begin{thebibliography}{10}
\providecommand{\url}[1]{#1}
\csname url@rmstyle\endcsname
\providecommand{\newblock}{\relax}
\providecommand{\bibinfo}[2]{#2}
\providecommand\BIBentrySTDinterwordspacing{\spaceskip=0pt\relax}
\providecommand\BIBentryALTinterwordstretchfactor{4}
\providecommand\BIBentryALTinterwordspacing{\spaceskip=\fontdimen2\font plus
\BIBentryALTinterwordstretchfactor\fontdimen3\font minus
  \fontdimen4\font\relax}
\providecommand\BIBforeignlanguage[2]{{%
\expandafter\ifx\csname l@#1\endcsname\relax
\typeout{** WARNING: IEEEtran.bst: No hyphenation pattern has been}%
\typeout{** loaded for the language `#1'. Using the pattern for}%
\typeout{** the default language instead.}%
\else
\language=\csname l@#1\endcsname
\fi
#2}}

\bibitem{beul2018fast}
M.~Beul, D.~Droeschel, M.~Nieuwenhuisen, J.~Quenzel, S.~Houben, and S.~Behnke,
  ``Fast autonomous flight in warehouses for inventory applications,''
  \emph{IEEE Robotics and Automation Letters (RA-L)}, vol.~3, no.~4, pp.
  3121--3128, 2018.

\bibitem{merino2012unmanned}
L.~Merino, F.~Caballero, J.~R. Mart{\'\i}nez-de Dios, I.~Maza, and A.~Ollero,
  ``An unmanned aircraft system for automatic forest fire monitoring and
  measurement,'' \emph{Journal of Intelligent \& Robotic Systems (JINT)},
  vol.~65, no.~1, pp. 533--548, 2012.

\bibitem{ravichandran2019uav}
R.~Ravichandran, D.~Ghose, and K.~Das, ``{UAV} based survivor search during
  floods,'' in \emph{International Conference on Unmanned Aircraft Systems
  (ICUAS)}.\hskip 1em plus 0.5em minus 0.4em\relax IEEE, 2019, pp. 1407--1415.

\bibitem{pecho2019unmanned}
P.~Pecho, P.~Magdolenov{\'a}, and M.~Bugaj, ``Unmanned aerial vehicle
  technology in the process of early fire localization of buildings,''
  \emph{Transportation {R}esearch {P}rocedia}, vol.~40, pp. 461--468, 2019.

\bibitem{mohta2018fast}
K.~Mohta, M.~Watterson, Y.~Mulgaonkar, S.~Liu, C.~Qu, A.~Makineni, K.~Saulnier,
  K.~Sun, A.~Zhu, J.~Delmerico, K.~Karydis, N.~Atanasov, G.~Loianno,
  D.~Scaramuzza, K.~Daniilidis, C.~J. Taylor, and V.~Kumar, ``Fast, autonomous
  flight in {GPS}-denied and cluttered environments,'' \emph{Journal of Field
  Robotics (JFR)}, vol.~35, no.~1, pp. 101--120, 2018.

\bibitem{mostafa2018radar}
M.~Mostafa, S.~Zahran, A.~Moussa, N.~El-Sheimy, and A.~Sesay, ``Radar and
  visual odometry integrated system aided navigation for {UAVS} in {GNSS}
  denied environment,'' \emph{Sensors}, vol.~18, no.~9, p. 2776, 2018.

\bibitem{spurny2021autonomous}
V.~Spurny, V.~Pritzl, V.~Walter, M.~Petrlik, T.~Baca, P.~Stepan, D.~Zaitlik,
  and M.~Saska, ``Autonomous firefighting inside buildings by an unmanned
  aerial vehicle,'' \emph{IEEE Access}, vol.~9, pp. 15\,872--15\,890, 2021.

\bibitem{mohamed2019survey}
S.~A. Mohamed, M.-H. Haghbayan, T.~Westerlund, J.~Heikkonen, H.~Tenhunen, and
  J.~Plosila, ``A survey on odometry for autonomous navigation systems,''
  \emph{IEEE Access}, vol.~7, pp. 97\,466--97\,486, 2019.

\bibitem{koyuncu2008probabilistic}
E.~Koyuncu and G.~Inalhan, ``A probabilistic {B}-spline motion planning
  algorithm for unmanned helicopters flying in dense {3D} environments,'' in
  \emph{IEEE/RSJ International Conference on Intelligent Robots and Systems},
  2008, pp. 815--821.

\bibitem{richter2016polynomial}
C.~Richter, A.~Bry, and N.~Roy, ``Polynomial trajectory planning for aggressive
  quadrotor flight in dense indoor environments,'' in \emph{Robotics
  Research}.\hskip 1em plus 0.5em minus 0.4em\relax Springer, 2016, pp.
  649--666.

\bibitem{nieuwenhuisen2019search}
M.~Nieuwenhuisen and S.~Behnke, ``Search-based {3D} planning and trajectory
  optimization for safe micro aerial vehicle flight under sensor visibility
  constraints,'' in \emph{IEEE International Conference on Robotics and
  Automation (ICRA)}, 2019, pp. 9123--9129.

\bibitem{liu2017search}
S.~Liu, N.~Atanasov, K.~Mohta, and V.~Kumar, ``Search-based motion planning for
  quadrotors using linear quadratic minimum time control,'' in \emph{IEEE/RSJ
  International Conference on Intelligent Robots and Systems (IROS)}, 2017, pp.
  2872--2879.

\bibitem{behnke2003local}
S.~Behnke, ``Local multiresolution path planning,'' in \emph{Robot Soccer World
  Cup}.\hskip 1em plus 0.5em minus 0.4em\relax Springer, 2003, pp. 332--343.

\bibitem{gonzalez2016adaptive}
A.~Gonz{\'a}lez-Sieira, M.~Mucientes, and A.~Bugar{\'\i}n, ``An adaptive
  multi-resolution state lattice approach for motion planning with
  uncertainty,'' in \emph{Robot 2015: Second Iberian Robotics
  Conference}.\hskip 1em plus 0.5em minus 0.4em\relax Springer, 2016, pp.
  257--268.

\bibitem{andersson2018receding}
O.~Andersson, O.~Ljungqvist, M.~Tiger, D.~Axehill, and F.~Heintz,
  ``Receding-horizon lattice-based motion planning with dynamic obstacle
  avoidance,'' in \emph{IEEE Conference on Decision and Control (CDC)}, 2018,
  pp. 4467--4474.

\bibitem{beul2020trajectory}
M.~Beul and S.~Behnke, ``Trajectory generation with fast lidar-based {3D}
  collision avoidance for agile {MAVs},'' in \emph{IEEE International Symposium
  on Safety, Security and Rescue Robotics (SSRR)}, 2020, pp. 42--48.

\bibitem{quenzel2021iros}
J.~Quenzel and S.~Behnke, ``Real-time multi-adaptive-resolution-surfel {6D
  LiDAR} odometry using continuous-time trajectory optimization,''
  \emph{arXiv:2105.02010}, 2021.

\bibitem{sommer2020cvpr}
C.~Sommer, V.~Usenko, D.~Schubert, N.~Demmel, and D.~Cremers, ``Efficient
  derivative computation for cumulative {B-Splines} on {Lie} groups,'' in
  \emph{IEEE Conference on Computer Vision and Pattern Recognition (CVPR)},
  2020.

\bibitem{amanatides1987fast}
J.~Amanatides and A.~Woo, ``A fast voxel traversal algorithm for ray tracing,''
  in \emph{Eurographics}, vol.~87, no.~3, 1987, pp. 3--10.

\bibitem{moore2014ekf}
T.~Moore and D.~Stouch, ``A generalized extended {Kalman} filter implementation
  for the robot operating system,'' in \emph{Int. Conf. on Intelligent
  Autonomous Systems (IAS)}.\hskip 1em plus 0.5em minus 0.4em\relax Springer,
  2014.

\bibitem{schleich2021search}
D.~Schleich and S.~Behnke, ``Search-based planning of dynamic {MAV}
  trajectories using local multiresolution state lattices,'' IEEE International
  Conference on Robotics and Automation (ICRA), 2021.

\bibitem{bentley1975multidimensional}
J.~L. Bentley, ``Multidimensional binary search trees used for associative
  searching,'' \emph{Communications of the ACM}, vol.~18, no.~9, pp. 509--517,
  1975.

\bibitem{beul2016icuas}
M.~Beul and S.~Behnke, ``Analytical time-optimal trajectory generation and
  control for multirotors,'' in \emph{International Conference on Unmanned
  Aircraft Systems (ICUAS)}, 2016.

\bibitem{beul2017icuas}
------, ``Fast full state trajectory generation for multirotors,'' in
  \emph{International Conference on Unmanned Aircraft Systems (ICUAS)}, 2017.

\end{thebibliography}
\vfill
\balance
\end{document}